\documentclass[
]{ceurart}

\usepackage{listings}
\usepackage[]{mdframed}
\usepackage{color}
\definecolor{mycolor}{gray}{0.80}
\usepackage[utf8]{inputenc}
\usepackage[T1]{fontenc}
\usepackage{listings}
\usepackage{placeins}

\begin{document}

\copyrightyear{2023}
\copyrightclause{Copyright for this paper by its authors.
  Use permitted under Creative Commons License Attribution 4.0
  International (CC BY 4.0).}

\conference{*version as accepted at the BioASQ Workshop at CLEF 2023}

\title{Is ChatGPT a Biomedical Expert?*}
\title[mode=sub]{Exploring the Zero-Shot Performance of Current GPT Models in Biomedical Tasks}


\author[1]{Samy Ateia}[%
email=Samy.Ateia@stud.uni-regensburg.de,
]
\author[1]{Udo Kruschwitz}[%
email=udo.kruschwitz@ur.de,
]
\address[1]{Information Science, University of Regensburg, Regensburg, Germany}


\begin{abstract}
We assessed the performance of commercial Large Language Models (LLMs) GPT-3.5-Turbo and GPT-4 on tasks from the 2023 BioASQ challenge. In Task 11b Phase B, which is focused on answer generation, both models demonstrated competitive abilities with leading systems. Remarkably, they achieved this with simple zero-shot learning, grounded with relevant snippets. Even without relevant snippets, their performance was decent, though not on par with the best systems. Interestingly, the older and cheaper GPT-3.5-Turbo system was able to compete with GPT-4 in the grounded Q\&A setting on Factoid and List answers. In Task 11b Phase A, focusing on retrieval, query expansion through zero-shot learning improved performance, but the models fell short compared to other systems. The code needed to rerun these experiments is available through GitHub\footnote{https://github.com/SamyAteia/bioasq}.

\end{abstract}

\begin{keywords}
  Zero-Shot Learning \sep
  LLMs \sep
  BioASQ \sep
  GPT-4 \sep
  NER \sep
  Question Answering
\end{keywords}

\maketitle

\section{Introduction}
Recently released ChatGPT models GPT-3.5-Turbo and GPT-4 \cite{openai2023gpt4} and their unprecedented zero-shot performance in a variety of tasks, sparked a surge in the development and application of LLMs. By participating in the eleventh CLEF BioASQ challenge \cite{BioASQ2023overview}, we wanted to explore how well these systems perform in specialized domains and whether they can compete with expert fine-tuned systems. 

\subsection{BioASQ Challenge}

BioASQ is a series of large-scale biomedical challenges associated with the CLEF 2023 conference. Its 11th iteration comprises three tasks \cite{BioASQ2023overview}:

\begin{enumerate}
    \item Synergy On Biomedical Semantic QA For Developing Issues
    \item Biomedical Semantic QA
    \item MedProcNER On MEDical PROCedure Named Entity Recognition
\end{enumerate}

This paper focuses on the second and third tasks, the two tasks we participated in. The Biomedical Semantic QA task (Task B) is subdivided into Phase A (document retrieval and snippet extraction) and Phase B (Question Answering) \cite{BioASQ2023task11bSynergy}. 

We will start with a brief overview of some related work in Section \ref{Related} before outlining the experimental setup in Section \ref{Setup}.  Section \ref{Methods} presents our methodology followed by a discussion of Results in Section \ref{Results}.  Finally, we will also touch on ethical issues (Section \ref{Ethics}) and offer some conclusions  (Section \ref{Conclusions}).

\section{Related Work}\label{Related}

To motivate our approach and contextualise our contribution  we will briefly discuss related work on recently released generative pre-trained transformer models, have a look at few-shot and zero-shot learning and touch on professional search, i.e. search in a professional context.

\subsection{GPT Models}
Recently released generative pre-trained transformer (GPT) models GPT-4 and GPT-3.5-turbo are based on the transformer architecture \cite{vaswani2017attention} and pre-trained on the next token prediction task. These models are additionally fine-tuned with reinforcement learning from human feedback, which greatly improves their ability to follow instructions and the perceived utility of their generations \cite{ouyang2022training}. OpenAI states that GPT-3.5-turbo is additionally optimized for chats, but does not disclose the exact training procedure used\footnote{https://platform.openai.com/docs/model-index-for-researchers}.

GPT-4 is the most recent and best performing model of OpenAI, which is, as of this writing, only programmatically accessible through closed beta API access\footnote{https://openai.com/product/gpt-4}. It exhibits human-level performance on various professional and academic benchmarks and can process images as well as text \cite{openai2023gpt4}.

\subsection{Few and Zero-Shot Learning}
These models improve over the earlier GPT-3 model which showed that in certain tasks sufficiently big LLMs can compete with fine-tuned transformer models using only few-shot learning, which greatly reduces the need for extensive training data \cite{brown2020language}.

In the \textbf{few-shot learning} setting, the GPT models are prompted with a text that contains a few examples of the tasks at hand, for example, multiple question-answer pairs, and at the end only the current question for which an answer should be generated by the model. The model then ideally completes this text by writing the correct answer.

In the \textbf{zero-shot learning} setting, the model is not supplied with any examples but rather only a direct question or abstract task description and is ideally able to generate a useful completion that answers the question or solves the task \cite{liu2023pre}.

Zero-shot and few-shot learning is especially interesting for applications in specialized domains with no or sparse training data available. Prior work in the biomedical domain has shown that language models pre-trained on in-domain data outperform models pre-trained on open domain data \cite{lee2020biobert}\cite{beltagy2019scibert}\cite{gu2021domain}. In this work, we want to explore whether these new GPT models, that are extensively trained on vast amounts of open domain data, can compete with specialized fine-tuned models that are expected to participate in the challenge. 

Even though these models are proprietary and neither the architecture nor the specific training process is known, several open-source alternatives have been developed such as OPT \cite{zhang2022opt}, BLOOM \cite{scao2022bloom}, or Pythia \cite{biderman2023pythia}. Projects based on these and other open source models are constantly improving, and some are already nearly reaching GPT-3.5-turbo level performance \cite{dettmers2023qlora}. We therefore believe that studying these commercial models is valuable for establishing a baseline in zero-shot performance for upcoming open-source alternatives. These alternatives could potentially challenge state-of-the-art (SOTA) systems across a wide range of natural language processing (NLP) tasks.

\subsection{Professional Search}

Professional search is search conducted in a work context \cite{Tait2014}. This is an everyday activity for many professionals that comes with specific 
 requirements which are different from the requirements of generic Web search \cite{DBLP:journals/sigir/VerberneHKWLRV18}. The BioASQ challenge can be framed as a form of professional search in which the searchers are  biomedical experts aiming to find answers to domain-specific questions. 

Automatic query expansion plays a key part in many professional search contexts including search by healthcare information professionals, patent agents and recruitment professionals \cite{doi:10.1177/02663821211034079} as well as 
in conducting systematic reviews \cite{MACFARLANE2022200091}. What is ultimately being submitted to the search system can turn out to be a fairly complex search strategy, a query involving domain-specific information based around Boolean operators. This is one of the motivations for us to explore automatic query expansion in our methodology.

\section{Experimental Setup}\label{Setup}

We describe the experimental setup of the two BioASQ tasks that we participated in, Task 11 B and MedProcNER. For Task 11 B a benchmark dataset with  training and test biomedical questions in English along with reference answers was used that has been created based on questions by biomedical experts \cite{krithara2023bioasq}.

\subsection{Task 11 B: Biomedical Semantic QA}

For \textbf{Phase A}, the participating systems receive a list of biomedical questions such as "\emph{Which protein is targeted by Herceptin?}" and should retrieve a list of up to 10 most relevant articles from the PubMed Annual Baseline Repository for 2023\footnote{https://lhncbc.nlm.nih.gov/ii/information/MBR.html}. Additionally, the systems should also create a list of at most 10 most relevant snippets extracted from the previously retrieved article titles or abstracts. Participating systems are compared based on the Mean Average Precision (MAP) metric.

In \textbf{Phase B}, the participating systems receive the same questions as in Phase A, along with a set of gold (correct) articles and snippets selected by biomedical experts. They should then generate an \emph{ideal} paragraph sized (at most 200 words) answer based on these snippets. The questions are also tagged with either \emph{Yes/No}, \emph{Factoid}, \emph{Summary}, or \emph{List} type indicating the format for an additional \emph{exact} answer that should be created by these systems.

\begin{itemize}
    \item \emph{Yes/no} questions require the exact answer to be either "yes" or "no".
    \item \emph{Factoid} question require the exact answer to be a list of up to 5 entity names or other short expressions ordered by decreasing confidence.
    \item \emph{List} questions require the exact answer to be a list of up to 100 entity names or similar short expressions. 
    \item \emph{Summary} questions do not require an additional exact answer, only the \emph{ideal} answer needs to be returned.
\end{itemize}

\subsection{MedProcNER: MEDical PROCedure Named Entity Recognition}

The \textbf{MedProcNER} task \cite{medprocner} focuses on the detection and mapping of medical procedures in Spanish texts. It consists of three subtasks:

\begin{itemize}
    \item In subtask 1, systems have to identify medical procedures from Spanish clinical reports.
    \item In subtask 2, systems have to map the medical procedures identified in subtask 1 to SNOMED CT codes \cite{stearns2001snomed}.
    \item In subtask 3, systems have to assign SNOMED CT codes to the full clinical report for later indexing.
\end{itemize}

\section{Methodology}\label{Methods}
\subsection{Model}
We accessed GPT-3.5-turbo and GPT-4 through the OpenAI API\footnote{https://platform.openai.com/docs/guides/chat/introduction}. We used a simple system message to set the behavior of the model, which can be seen in Listing \ref{systemListing}.

\begin{lstlisting}[frame=single,caption={System Message},label=systemListing,breakindent=0pt,columns=fullflexible]
You are BioASQ-GPT, an AI expert in question answering, research, and information retrieval in the biomedical domain.
\end{lstlisting}

This system message was then followed by task specific zero-shot prompts, including necessary information such as the questions, snippets, or retrieved article titles. More details on these prompts can be found in the subsection corresponding to the particular task. Prompt engineering has developed into a very active field and at this point we should note that there is scope for plenty of future work exploring more systematically the best way of prompting the system for the task at hand.

We experimented with a subset of the BioASQ training and development data to explore the system's behavior and evaluate the performance of individual modules. 

Additional parameters that were sent in the API request to the models were \emph{temperature} which controls the randomness of completion; \emph{frequency\_penalty} which discourages repetition of words or phrases; and \emph{presence\_penalty} which has a similar effect. We set \emph{temperature} to 0 for all requests to have reproducible results over multiple runs. 

As these models are currently \textbf{non-deterministic}, even with temperature set to 0, there is a residual randomness in the generations, which can lead to slightly different results in each run\footnote{https://platform.openai.com/docs/models/gpt-3-5}. We also conducted a limited test to roughly estimate the variance of the results by repeating five runs over the same 50 questions.

\subsection{Task 11 B}
\subsubsection{Phase A}
Our approach used zero-shot learning for query expansion, query reformulation and reranking directly with the models. For document retrieval, we queried the eUtils API with a \emph{maxdate} cutoff corresponding to the creation date of the relevant 2023 PubMed snapshot. The Entrez Programming Utilities (eUtils) API is a set of web applications provided by the National Center for Biotechnology Information (NCBI), which offers programmatic access to the various databases and functionalities of the NCBI resources, such as PubMed. We also used the sort by relevance option of PubMed and retrieved only the top 50 results for a given query. 

We acknowledge that querying the live PubMed database with the corresponding date cutoff is not the same as searching through the downloaded static snapshot or using the search interface provided by the BioASQ organizers. Articles could be deleted or modified in PubMed, which could affect the reproducibility and comparability of the results with other systems. To estimate the impact of this approach, we looked up all articles that were included in the gold set provided in Phase B of the task after the challenge concluded and found that one out of the 899 referenced articles was no longer retrievable in PubMed\footnote{Article from batch 4 that is no longer accessible in PubMed: http://www.ncbi.nlm.nih.gov/pubmed/36459075}.

We were most interested in the impact of the query expansion step and therefore conducted one run with and one without query expansion for both models, where we instead sent the question directly as a query to PubMed.

The exact steps were:
\begin{enumerate}
\item Query expansion
\item Search on PubMed
\item Query refinement only if no documents were found and one additional search on PubMed
\item Reranking of top 50 articles based on title string
\end{enumerate}
All of these steps were executed automatically in Python without manual intervention, the exact code used is available on GitHub\footnote{https://github.com/SamyAteia/bioasq}.
The zero-shot learning prompt used for query expansion can be seen in Listing \ref{expansionListing}. Where the placeholder \emph{\{question\}} was replaced by the question that was currently processed by the system. For query expansion, we set \emph{frequency\_penalty} to 0.5 and \emph{presence\_penalty} to 0.1.

\begin{lstlisting}[float,language=python,frame=single,caption={Query Expansion Prompt},label=expansionListing,breakindent=0pt,columns=fullflexible]
{"role": "user", "content": f"""Expand this search query:
'{question}' for PubMed by incorporating synonyms and additional terms that closely relate to the main topic and help reduce ambiguity. Assume that phrases are not stemmed; therefore, generate useful variations. Return only the query that can directly be used without any explanation text. Focus on maintaining the query's precision and relevance to the original question."""}
\end{lstlisting}

Some example query expansions for this prompt can be seen in Listing \ref{expansionExample}. Interestingly, these models seem to not only know what Boolean syntax is accepted by PubMed but also important internal fields such as \emph{MeSH Terms} and the syntax on how to query on these fields, but these were not often used in the expanded queries.\footnote{The identification of suitable MeSH terms in structured queries for systematic reviews has been explored in detail elsewhere, e.g. \cite{10.1145/3503516.3503530,10.1145/3539597.3573025}}

\begin{lstlisting}[frame=single,caption={Query Expansion Example},label=expansionExample,breakindent=0pt,columns=fullflexible]
Question: What are the outcomes of ubiquitination?

Expanded Query: ("ubiquitination" OR "ubiquitin modification" OR "ubiquitin conjugation" OR "ubiquitin pathway") AND ("outcomes" OR "effects" OR "consequences")

Question: What is the incidence of Leigh syndrome?

Expanded Query: ("Leigh syndrome"[MeSH Terms] OR "Leigh syndrome"[All Fields] OR "subacute necrotizing encephalomyelopathy"[All Fields]) AND ("incidence"[MeSH Terms] OR "incidence"[All Fields] OR "prevalence"[MeSH Terms] OR "prevalence"[All Fields])
\end{lstlisting}

For the optional query reformulation step, we used the prompt in Listing \ref{refinementPrompt}. This step was introduced after it became clear that some queries constructed by the models were overly specific and returned no results. The placeholder \emph{\{question\}} in the prompt was replaced by the question that was currently processed by the system, and the placeholder \emph{\{original\_query\}} was replaced by the original expanded query that returned no results. For query reformulation, we set \emph{frequency\_penalty} to 0.6 and \emph{presence\_penalty} to 0.2.
An example of a query reformulation that generated a slightly broader query that then led to some results can be seen in Listing \ref{reformulationExample}. Additionally, terms added to the query are highlighted in gray.

\begin{lstlisting}[frame=single,caption={Query Reformulation Prompt},label=refinementPrompt,breakindent=0pt,columns=fullflexible]
{"role": "user", "content": f"""Given that the following search query for PubMed has returned
no documents, please generate a broader query that retains the original question's context and relevance. Assume that phrases are not stemmed; therefore, generate useful variations. Return only the query that can directly be used without any explanation text. Focus on maintaining the query's precision and relevance to the original question. Original question: '{question}', Original query: '{original_query}'."""}
\end{lstlisting}

\begin{lstlisting}[float,frame=single,caption={Query Reformulation Example},label=reformulationExample,breakindent=0pt,columns=fullflexible, escapeinside={(*}{*)}]
Question: Can skin picking phenotype present following methylphenidate treatment?
Query: ("skin picking" OR "excoriation disorder" OR "dermatillomania" OR "compulsive skin picking") AND (phenotype OR presentation OR manifestation) AND ("methylphenidate treatment" OR "methylphenidate therapy" OR "methylphenidate administration")

Reformulated Query: ("skin picking" OR "excoriation disorder" OR "dermatillomania" OR "compulsive skin picking") AND (phenotype OR presentation OR manifestation OR (*\colorbox{mycolor}{symptoms}*)) AND ("methylphenidate treatment" OR "methylphenidate therapy" OR "methylphenidate administration" OR (*\colorbox{mycolor}{"methylphenidate use"}*))
\end{lstlisting}

For the final reranking step, we took the titles of the top 50 returned articles as returned by the relevancy sort from PubMed and prompted the model to rerank these articles given the original question and return the top 10 articles. The prompt used for the reranking can be seen in Listing \ref{rerankingPrompt}, where \emph{\{articles\_str\}} is replaced by the list of returned article titles, \emph{\{question\}} is replaced by the question that is currently processed by the system, and \emph{\{nr\_of\_articles\}} is replaced by the 10 or fewer articles if less relevant articles were returned by PubMed. For reranking, we set \emph{frequency\_penalty} to 0.3 and \emph{presence\_penalty} to 0.1.

\begin{lstlisting}[frame=single,caption={Reranking Prompt},label=rerankingPrompt,breakindent=0pt,columns=fullflexible]
{"role": "user", "content": f"{articles_str} \n\n Given these articles and the question: '{question}'. Rerank the articles based on their relevance to the question and return the top {nr_of_articles} most relevant articles as a comma separated list of their index ids. Don't explain your answer, return only this list, for example: '1, 2, 3, 4' "}
\end{lstlisting}

The returned list was then mapped back to the articles retrieved from PubMed, and these were returned as the required output of Phase A.

We also explored the extraction of snippets for the Phase A task but abandoned it, as it required sending all abstracts of the 10 returned papers for processing to the model, which was especially expensive for the GPT-4 model because API usage is priced on token counts, and we were exploring these models on a limited budget.

\subsubsection{Phase B}
In Phase B, we used the gold (correct) snippets from the test set and sent them along with the question and description of the answer format to the model. 

We also conducted a test where this grounding information in the form of relevant snippets was omitted and just the question and description of the answer format were sent to the models.

The prompts utilized for generating these answer types are listed as follows: for ideal answers, refer to Listing \ref{idealAnswer}; for Yes/No answers, see Listing \ref{yesnoPrompt}; for List answers, Listing \ref{factoidPrompt}; and for Factoid responses, see Listing \ref{listPrompt}.

In all these prompts, \emph{\{question['body']\}} is replaced by the question that is currently processed by the system, and \emph{\{snippets\}} is replaced by the snippets provided by the test set.

For all answer types, we set \emph{frequency\_penalty} to 0.5. \emph{Presence\_penalty} was set to 0.3 for Yes/No answers, to 0.1 for both List and Factoid answers, and to 0.7 for the ideal answer type.

\begin{lstlisting}[float,frame=single,caption={Ideal Answer Prompt},label=idealAnswer,breakindent=0pt,columns=fullflexible]
{"role": "user", "content": f""" {snippets}\n\n\ '{question['body']}'. Answer this question by returning a single paragraph-sized text ideally summarizing the most relevant information. The maximum allowed length of the answer is 200 words. The returned answer is intended to approximate a short text that a biomedical expert would write to answer the corresponding question (e.g., including prominent supportive information)."""}
\end{lstlisting}

\begin{lstlisting}[float,frame=single,caption={Yes/No Answer Prompt},label=yesnoPrompt,breakindent=0pt,columns=fullflexible]
{"role": "user", "content": f" {snippets}\n\n\ '{question['body']}'. You *must answer* only with lowercase 'yes' or 'no' even if you are not sure about the answer."}
\end{lstlisting}

\begin{lstlisting}[float,frame=single,caption={Factoid Answer Prompt},label=factoidPrompt,breakindent=0pt,columns=fullflexible]
 {"role": "user", "content": f" {snippets}\n\n\ '{question['body']}'. Answer this question by returning only a JSON string array of entity names, numbers, or similar short expressions that are an answer to the question, ordered by decreasing confidence. The array should contain at max 5 elements but can contain less. If you don't know any answer return an empty list. Return only this list, it must not contain phrases and **must be valid JSON**."}
\end{lstlisting}

\begin{lstlisting}[float,frame=single,caption={List Answer Prompt},label=listPrompt,breakindent=0pt,columns=fullflexible]
{"role": "user", "content": f" {snippets}\n\n\ '{question['body']}'. Answer this question by only returning a JSON string array of entity names, numbers, or similar short expressions that are an answer to the question (e.g., the most common symptoms of a disease). The returned list will have to contain no more than 100 entries of no more than 100 characters each. If you don't know any answer return an empty list. Return only this list, it must not contain phrases and **must be valid JSON**."}
\end{lstlisting}

\subsection{MedProcNER}
For the MedProcNER task, we translated all prompt templates, including the system prompt, to Spanish using and comparing deepL\footnote{https://www.deepl.com/en/translator}) and ChatGPT\footnote{https://chat.openai.com/}. For substask 1, instead of using zero-shot prompting as before, we instead explored the few-shot prompting approach, where we included three examples from the training set into the request sent to the OpenAI API. We also compared the performance of GPT-3.5-turbo and GPT-4.

The relevant Python code part that constructed the prompt can be seen in Listing \ref{medprocPrompt}. The \emph{examples} list mentioned therein contained three examples taken from the training set. 
\begin{lstlisting}[float,frame=single,caption={MedProcNER Prompt},label=medprocPrompt,breakindent=0pt,columns=fullflexible]
conversation = [{'role': 'system', 'content': """Eres un asistente útil que extrae procedimientos médicos de textos médicos en español. Un procedimiento médico se refiere a cualquier acción diagnóstica, terapéutica, médica o quirúrgica realizada en un paciente. Tu respuesta debe ser una lista de procedimientos en formato JSON válido."""}]
for input, output in examples:
    conversation.append({'role': 'user', 'content': f'{input}'})
    conversation.append({'role': 'assistant', 'content': json.dumps(output)})
conversation.append({'role': 'user', 'content': f"""Extraiga todos los procedimientos médicos del texto delimitado por tres comillas invertidas. Devuelve una lista vacía si no se menciona ninguno. {text}"""})
\end{lstlisting}

For substask 2, we used the gazetteer file provided by the MedProcNER task organizers. We filtered the file for all SNOMED CT codes that were tagged as  procedure and stemmed their terms, and used Levenshtein distance based fuzzy matching to find an entry for a procedure. The detailed code used for all tasks is available on the aforementioned GitHub repository.

For subtask 3, we just joined all SNOMED CT codes identified in subtask 2 for one document. 

\section{Results}\label{Results}
The systems participating in the Biomedical Semantic Q\&A task were evaluated in four batches.
Results are reported for every batch. For readability, we only included the results of our systems and the top performing systems. The full result tables are publicly available on the BioASQ website\footnote{http://participants-area.bioasq.org/results/11b/phaseA/}
\subsection{Task 11 B Phase A}
We participated with 4 systems in Task 11 B Phase A, the systems' names and their properties are listed as follows:
\begin{itemize}
    \item UR-gpt4-zero-ret corresponds to GPT-4 with query expansion.
    \item UR-gpt3.5-turbo-zero corresponds to GPT-3.5-turbo with query expansion.
    \item UR-gpt4-simple corresponds to GTP-4 without query expansion.
    \item UR-gpt3.5-t-simple corresponds to GPT-3.5-turbo without query expansion.
\end{itemize}

The following Table \ref{tab:phaseABatches} shows the results of our systems participating in the 4 batches. MAP was the official metric to compare the systems. N stands for the number of participating systems in each batch.

\begin{table}[h!]
\caption{Task 11 B Phase A, Batches 1-4}
\footnotesize
\label{tab:phaseABatches}
\begin{tabular}{|c|c|c|c|c|c|c|c|}
\hline
Batch & Position & System & Precision & Recall & F-Measure & \textbf{MAP} & GMAP \\
\hline
 & 1 & Top Competitor & 0.2118 & 0.6047 & 0.2774 & 0.4590 & 0.0267 \\
 & 19 & UR-gpt4-zero-ret & 0.1664 & 0.3352 & 0.1955 & 0.2657 & 0.0009 \\
Batch 1 & 21 & UR-gpt3.5-turbo-zero & 0.1488 & 0.2847 & 0.1782 & 0.2145 & 0.0009 \\
N = 33 & 24 & UR-gpt4-simple & 0.1654 & 0.2508 & 0.1799 & 0.1809 & 0.0005 \\
 & 25 & UR-gpt3.5-t-simple & 0.1600 & 0.2290 & 0.1734 & 0.1769 & 0.0003 \\
\hline
 & 1 & Top Competitor & 0.1027 & 0.5149 & 0.1618 & 0.3852 & 0.0104 \\
Batch 2 & 20 & UR-gpt4-simple & 0.0945 & 0.3011 & 0.1277 & 0.1905 & 0.0011 \\
N = 33 & 21 & UR-gpt3.5-turbo-zero & 0.1153 & 0.2977 & 0.1455 & 0.1736 & 0.0008 \\
\hline
 & 1 & Top Competitor & 0.0800 & 0.4776 & 0.1320 & 0.3185 & 0.0049 \\
 & 21 & UR-gpt3.5-turbo-zero & 0.1295 & 0.3258 & 0.1646 & 0.2048 & 0.0008 \\
Batch 3 & 22 & UR-gpt4-zero-ret & 0.1086 & 0.2289 & 0.1303 & 0.1930 & 0.0003 \\
N = 35 & 23 & UR-gpt4-simple & 0.1089 & 0.2102 & 0.1238 & 0.1727 & 0.0002 \\
 & 24 & UR-gpt3.5-t-simple & 0.1078 & 0.1981 & 0.1217 & 0.1553 & 0.0002 \\
\hline
 & 1 & Top Competitor & 0.0933 & 0.4292 & 0.1425 & 0.3224 & 0.0030 \\
 & 18 & UR-gpt4-zero-ret & 0.0791 & 0.1728 & 0.0933 & 0.1251 & 0.0002 \\
Batch 4 & 19 & UR-gpt3.5-turbo-zero & 0.0922 & 0.1956 & 0.1025 & 0.1139 & 0.0002 \\
N = 27 & 20 & UR-gpt4-simple & 0.0785 & 0.1563 & 0.0864 & 0.1010 & 0.0002 \\
 & 21 & UR-gpt3.5-t-simple & 0.0752 & 0.1319 & 0.0810 & 0.0912 & 0.0001 \\
\hline
\end{tabular}
\end{table}

One observation is that GPT-4 achieved better results than GPT-3.5-turbo in all batches except batch 3. It seems to perform better in both query expansion and reranking without query expansion. Query expansion consistently improves the results for all models in all batches. It greatly improves recall in all batches, and in most batches, precision is also slightly increased except in batch 1, where it leads to decreased precision for GPT-3.5-turbo but an overall improved F1 score.

In general, our approach performs worse than most systems. This could be due to the fact that we do not do any embedding based neural retrieval, but instead only rely on the keywords created by the models in the query expansion step and the relevancy ranking provided by PubMed. The reranking window of only 50 article titles might also be too small, or the information provided by the titles is not sufficient for a more effective reranking. A thorough ablation study in future work could help explain the contribution of these individual factors to the overall system performance.

Using only query expansion in the retrieval phase and not having to do any embedding calculations during indexing does come with advantages for applying such an approach to existing or huge search use-cases where efficient reindexing with more advanced embedding based approaches might not be feasible. On the other hand, the used models do take several seconds to create results for both reranking and query expansion, which could limit their usefulness in classical enterprise-search use-cases if sub-second response times are expected.

\subsection{Task 11 B Phase A}
We participated with 4 systems in Task 11 B Phase B, the systems' names and their properties are listed as follows:
\begin{itemize}
    \item \emph{UR-gpt4-zero-ret} corresponds to GPT-4 grounded with snippets.
    \item \emph{UR-gpt3.5-turbo-zero} corresponds to GPT-3.5-turbo grounded with snippets.
    \item \emph{UR-gpt4-simple} corresponds to GTP-4 answering directly without reading snippets.
    \item \emph{UR-gpt3.5-t-simple} corresponds to GPT-3.5-turbo answering directly without reading snippets.
\end{itemize}

We were not able to complete all runs in batches 1 and 2, which is why some results are missing. We report the results for each answer format (Yes/No, Factoid, List) separately in the following tables. For readability, we again only included the results of our systems and the top-performing systems, the full result tables are publicly available on the BioASQ website\footnote{http://participants-area.bioasq.org/results/11b/phaseB/}.

\begin{table}[h!]
\caption{Task 11 B Phase B, Yes/No Questions Batches 1-4}
\footnotesize
\label{tab:phaseBYesNoBatches}
\begin{tabular}{|c|c|c|c|c|c|c|}
\hline
Batch & Position & System & Accuracy & F1 Yes & F1 No & \textbf{Macro F1} \\
\hline
 & 1 & Top Competitor & 0.9583 & 0.9697 & 0.9333 & 0.9515 \\
Batch1 & 8 & UR-gpt4-zero-ret & 0.9167 & 0.9412 & 0.8571 & 0.8992 \\
N = 33 & 9 & UR-gpt4-simple & 0.9167 & 0.9412 & 0.8571 & 0.8992 \\
 & 13 & UR-gpt3.5-turbo-zero & 0.8750 & 0.9091 & 0.8000 & 0.8545 \\
\hline
& 1 & Top Competitor & 1.0000 & 1.0000 & 1.0000 & 1.0000 \\
Batch2 & 7 & UR-gpt4-zero-ret & 0.9583 & 0.9655 & 0.9474 & 0.9564 \\
N = 42  & 12 & UR-gpt3.5-turbo-zero & 0.9167 & 0.9333 & 0.8889 & 0.9111 \\
\hline
 & 1 & Top Competitor & 1.0000 & 1.0000 & 1.0000 & 1.0000 \\
 & 9 & UR-gpt4-zero-ret & 0.9167 & 0.9375 & 0.8750 & 0.9063 \\
Batch3 & 12 & UR-gpt4-simple & 0.8750 & 0.9032 & 0.8235 & 0.8634 \\
N = 47 & 14 & UR-gpt3.5-turbo-zero & 0.8750 & 0.9091 & 0.8000 & 0.8545 \\
 & 21 & UR-gpt3.5-t-simple & 0.7917 & 0.8485 & 0.6667 & 0.7576 \\
\hline
 & 1 & Top Competitor & 1.0000 & 1.0000 & 1.0000 & 1.0000 \\
 & 7 & UR-gpt4-zero-ret & 0.9286 & 0.8889 & 0.9474 & 0.9181 \\
Batch4 & 14 & UR-gpt3.5-turbo-zero & 0.9286 & 0.8571 & 0.9524 & 0.9048 \\
N = 52 & 19 & UR-gpt4-simple & 0.7857 & 0.7273 & 0.8235 & 0.7754 \\
 & 29 & UR-gpt3.5-t-simple & 0.4286 & 0.5000 & 0.3333 & 0.4167 \\
\hline
\end{tabular}
\end{table}

\begin{table}[h!]
\caption{Task 11 B Phase B, Factoid Questions Batches 1-4}
\footnotesize
\label{tab:phaseBFactoidBatches}
\begin{tabular}{|c|c|c|c|c|c|}
\hline
Batch & Position & System & Strict Acc. & Lenient Acc. & \textbf{MRR} \\
\hline
 & \textbf{1} & UR-gpt4-zero-ret & 0.5789 & 0.5789 & 0.5789 \\
Batch1 & \textbf{2} & UR-gpt3.5-turbo-zero & 0.5263 & 0.6316 & 0.5789 \\
N = 33  & 3 & Next Competitor & 0.5263 & 0.6316 & 0.5570 \\
 & 22 & UR-gpt4-simple & 0.2105 & 0.2632 & 0.2368 \\
\hline
 & 1 & Top Competitor & 0.5455 & 0.6364 & 0.5909 \\
Batch2 & 2 & Next Competitor & 0.5455 & 0.6364 & 0.5909 \\
N = 42 & 3 & UR-gpt3.5-turbo-zero & 0.5455 & 0.5909 & 0.5682 \\
 & 4 & UR-gpt4-zero-ret & 0.5455 & 0.5909 & 0.5682 \\
\hline
& 1 & Top Competitor & 0.4615 & 0.6538 & 0.5205 \\
& 5 & UR-gpt3.5-turbo-zero & 0.5000 & 0.5000 & 0.5000 \\
Batch3 & 11 & UR-gpt4-zero-ret & 0.4615 & 0.5000 & 0.4808 \\
N = 47 & 22 & UR-gpt4-simple & 0.2692 & 0.4615 & 0.3654 \\
 & 27 & UR-gpt3.5-t-simple & 0.3077 & 0.3077 & 0.3077 \\
\hline
 & 1 & Top Competitor & 0.6452 & 0.8710 & 0.7323 \\
Batch4 & 6 & UR-gpt3.5-turbo-zero & 0.6452 & 0.6452 & 0.6452 \\
N = 52 & 13 & UR-gpt4-zero-ret & 0.5161 & 0.6129 & 0.5645 \\
 & 30 & UR-gpt3.5-t-simple & 0.2581 & 0.2903 & 0.2742 \\
 & 33 & UR-gpt4-simple & 0.2258 & 0.2581 & 0.2366 \\
 \hline
\end{tabular}
\end{table}

\begin{table}[h!]
\caption{Task 11 B Phase B, List Questions Batches 1-4}
\footnotesize
\label{tab:phaseBListBatches}
\begin{tabular}{|c|c|c|c|c|c|}
\hline
Batch & Position & System & Strict Acc. & Lenient Acc. & \textbf{MRR} \\
\hline
 & 1 & Top Competitor & 0.7861 & 0.6668 & 0.7027 \\
Batch1 & \textbf{2} & UR-gpt3.5-turbo-zero & 0.6742 & 0.7249 & 0.6917 \\
N = 33  & 8 & UR-gpt4-zero-ret & 0.6472 & 0.6530 & 0.6495 \\
 & 19 & UR-gpt4-simple & 0.4000 & 0.4014 & 0.3939 \\
\hline
 & \textbf{1} & UR-gpt3.5-turbo-zero & 0.4598 & 0.4671 & 0.4316 \\
Batch2 & 2 & Next Competitor & 0.5099 & 0.3577 & 0.3980 \\
N = 42 & 4 & UR-gpt4-zero-ret & 0.3742 & 0.4369 & 0.3828 \\
\hline
& 1 & Top Competitor & 0.6519 & 0.6058 & 0.6049 \\
& 3 & UR-gpt4-zero-ret & 0.5518 & 0.6597 & 0.5736 \\
Batch3 & 9 & UR-gpt3.5-turbo-zero & 0.5600 & 0.5140 & 0.5101 \\
N = 47 & 24 & UR-gpt3.5-t-simple & 0.2690 & 0.2385 & 0.2333 \\
 & 25 & UR-gpt4-simple & 0.2519 & 0.2343 & 0.2305 \\
\hline
 & 1 & Top Competitor & 0.7139 & 0.8061 & 0.7440 \\
Batch4 & 2 & UR-gpt4-zero-ret & 0.6902 & 0.7818 & 0.7191 \\
N = 52 & 10 & UR-gpt3.5-turbo-zero & 0.6090 & 0.6710 & 0.6196 \\
 & 21 & UR-gpt4-simple & 0.4440 & 0.4214 & 0.4127 \\
 & 26 & UR-gpt3.5-t-simple & 0.3944 & 0.3362 & 0.3470 \\
 \hline
\end{tabular}
\end{table}

In the Yes/No question format, our results indicate that GPT-4 surpasses GPT-3.5-turbo in both the grounded and ungrounded settings. For batches 1 and 3, the ungrounded GPT-4 system \emph{UR-gpt4-simple} even showed a tendency to perform better than the grounded variant of GPT-3.5-turbo \emph{UR-gpt3.5-turbo-zero} as can be seen in Table \ref{tab:phaseBYesNoBatches}.

In the Factoid question format, both grounded GPT-4 and grounded GPT-3.5-turbo achieved an MRR score of 0.5789 taking first and second place over all other systems. In the remaining batches, GPT-3.5-turbo stayed consistently in the top 6 systems, while GPT-4 only reached 11th and 13th place in batches 3 and 4. This mixed performance comparison between GPT-3.5-turbo and GPT-4 was also observed in the List question format, where GPT-3.5-turbo achieved 1st place in batch 2 but was behind GPT-4 in batches 3 and 4. The results for the Factoid question format are shown in Table \ref{tab:phaseBFactoidBatches} and the results for the List question format are shown in Table \ref{tab:phaseBListBatches}.

While GPT-4 seems to perform consistently better than GPT-3.5-turbo in the Yes/No question format, there is no clear winner in the more extractive Factoid and List formats.

Both models without grounding information from snippets were not able to compete with the top models but were often placed slightly below the average performing systems, which is still surprisingly good as in this setting the models need to rely only on the open-domain knowledge acquired during training for answering these questions.

\subsection{Task MedProcNER}
In the MedProcNER task, GPT-4 performed better than GPT-3.5-turbo, but  was not able to compete with the best performing system. The results are shown in Table \ref{tab:medProcNer}. Our simple gazetteer based entity linking and indexing approach performed poorly compared to the top-performing system. At the time of this writing, the performance of other systems involved in the task has not been published yet.

\begin{table}[h]
\caption{Comparison of F1 scores of different systems for NER, Entity Linking, and Indexing tasks.}
\label{tab:medProcNer}
\begin{tabular}{|c|c|c|c|}
\hline
\textbf{Task} & \textbf{Top Performing System F1} & \textbf{GPT-3.5-turbo F1} & \textbf{GPT-4 F1}\\
\hline
NER & 0.7985 & 0.3002 & 0.4814\\
\hline
EL & 0.5707 & 0.1264 & 0.1976\\
\hline
Indexing & 0.6242 & 0.1785 & 0.2695\\
\hline
\end{tabular}
\end{table}

Even though the few-shot NER approach did not compete with the top-performing system in the MedProcNER task, it still indicates that GPT-4 can be used for specialized domains in multilingual tasks while only using a minimal amount of training data.

\subsection{Discussion and Future Work}
The results from our participation in the BioASQ challenge indicate that current commercial GPT models GPT-3.5-turbo and GPT-4 can compete with other presumably fine-tuned leading systems in question answering in the biomedical domain, while only being zero-shot prompted with relevant snippets. Even without relevant snippets, just relying on the biomedical knowledge aquired during their pre-training, these models were performing better than some of the other systems participating in the task.

One big challenge in using zero-shot learning with these GPT models is prompt-engineering. It still seems to be more of an art than a science and requires considerable testing \cite{zhou2023large}. During system development, it became clear that the expanded queries in Task 11 B Phase A were sometimes too specific and did not return results. We tried to prompt the models to create broader queries that were not using as many phrase terms that are not stemmed in PubMed, but the overall system performance on our development set declined. We therefore experimented with using GPT-4 to come up with a better prompt by supplying it with the original prompt and the 5 worst-performing and 5 best-performing queries. The new prompt actually increased the performance of the system. This self prompt learning might be an interesting approach to investigate further in future work.

Nevertheless, the zero-shot learning approach makes the usage of these models very accessible, as it does not require thorough data preparation, knowledge about classical deep learning techniques, or advanced programming skills.

A prominent problem in these GPT models are so-called hallucinations \cite{ji2023survey}. These are unsupported or factually wrong statements in the responses. These problems might be especially observable in the ideal answer setting. In future work, we want to conduct a thorough investigation of the factuality of the ideal answers and especially compare the grounded and ungrounded settings. This could provide error rate estimates that might be useful for generative search systems in specialized domains. 

As noted earlier, these commercial models are not completely deterministic, even when the temperature parameter is set to 0. OpenAI states in their documentation: 
\begin{quote}
    "OpenAI models are non-deterministic, meaning that identical inputs can yield different outputs. Setting the temperature parameter to 0 will make the outputs mostly deterministic, but a small amount of variability may remain."\footnote{https://platform.openai.com/docs/models/gpt-3-5}
\end{quote}
We had concerns about the potential cascading effect of such residual non-determinism, especially in the context of query expansion. To estimate this variability, we performed a limited test by repeating the retrieval task from Task 11 B Phase A over 50 questions taken from the training set five times with the same model. Our test results showed minimal variance across metrics such as MAP, precision, recall, and F-measure, indicating that while variability exists, its impact is currently minimal, with broader investigations pending for future work.

This residual non-determinism in the model output also led to some instability in the system when we fully relied on the model returning the right output format for further processing. For example, in the Yes/No question format, the evaluation system of the BioASQ organizers expects the answers to be all lowercase, either "yes" or "no". The models often returned variants such as "Yes" or "Yes." even if explicitly prompted not to do so. This necessitated an additional normalization post-processing step.

In the MedProcNER task, where we used few-shot learning, it seemed that the examples greatly assisted the model in returning the correct output format. We suspect that giving even just a few examples is a more effective way to guide the models towards the expected output format than explicitly describing the format in a zero-shot learning prompt.

Even if the models were outputting the right format, the overall system was still unstable due to the instability of the OpenAI API. In every run, there were at least 2–3 requests that failed due to internal server errors or the model being overloaded with requests. Thus, retry loops must be incorporated when accessing such external services.

As usage of these models is priced based on token count, some use-cases might not be financially feasible yet. Only running one evaluation batch with GPT-4 can cost around \$10 in model usage. At the same time, the GPT-4 model was still much slower in answering requests than GPT-3.5-turbo. These two factors led us to not participate in the snippet generation task, as this task is especially demanding regarding both the amount of tokens to be processed in the prompt and generated as a response. In general, the economic barrier to using these commercial models may hinder some researchers due to the cost of usage. Also, over-reliance on these models might stifle innovation in other research areas.

We also conducted a limited test with grounding the query expansion by suggesting semantically related terms from the word embeddings supplied by the BioASQ organizers, but these terms led to queries that performed worse than just ungrounded ones. We did not investigate this approach thoroughly and leave it open for future work.

Some of our results might indicate that the performance gap between presumably smaller (GPT-3.5-turbo) and more complex models (GPT-4) is narrower in the grounded extractive Q\&A setting, because GPT-3.5-turbo sometimes performed better than GPT-4 in answering Factoid or List questions in some of the batches. It would be interesting to see how model performance in this setting scales with model size, and to test whether the use of much smaller generative models is feasible. Some related work in other use-cases already showed promising results in this direction \cite{eldan2023tinystories}\cite{ho2023large}. This might open up new possibilities for using these models in enterprise search settings where confidential data must remain on-premise \cite{kruschwitz2017searching}.

\section{Ethical Considerations}\label{Ethics}

The use of large language models like GPT-3.5-Turbo and GPT-4 in biomedical tasks presents several ethical considerations. 

First, we must address data privacy. While these models do not retain specific training examples, there is a remote possibility of them generating outputs resembling sensitive data, or sensitive data included in a prompt might be repeated and further processed in downstream tasks. This issue has to be addressed when employing these models in a real world biomedical context.

Second, as these models may produce factually incorrect outputs or "hallucinations" \cite{ji2023survey}, rigorous fact-checking mechanisms must be applied, especially when used in a biomedical context to prevent the spread of harmful misinformation.

Lastly, large language models operate as black-box algorithms, raising issues of interpretability, transparency, and accountability \cite{bender2021dangers}. 

In conclusion, the potential of large language models in biomedical tasks is significant, but the ethical implications of their deployment need careful attention.

\section{Conclusion}\label{Conclusions}
We showed that in context learning, both zero- and few-shot, with recent LLMs trained on human feedback can compete with presumably fine-tuned state-of-the-art systems in some domain-specific questions answering tasks. Zero- and few-shot learning can greatly simplify and speed up the development of complex NLP or IR systems, which might be especially useful for research and prototyping. It also opens up the possibility to improve use-cases where fine-tuning is not feasible due to a lack of available training data. 

Prompt engineering for these models poses challenges, and grounding the answer generation with the right context information is an interesting problem for current and future generative search systems research. Even though the currently offered GPT models have severe limitations regarding cost of usage, speed, and factuality, we see promising research towards making these types of models more affordable and accessible and improving their overall performance and factuality.

\begin{acknowledgments}
  We want to thank the organizers of the BioASQ challenge for setting up this challenge and supporting us during our participation. We are also grateful for the feedback and recommendations of the anonymous reviewers.
\end{acknowledgments}

\FloatBarrier
\bibliography{bibliography}

\appendix

\end{document}